\documentclass{vgtc}                          

\ifpdf
  \pdfoutput=1\relax                   
  \usepackage{graphicx}                
  \DeclareGraphicsExtensions{.pdf,.png,.jpg,.jpeg} 
\else
  \usepackage{graphicx}                
  \DeclareGraphicsExtensions{.eps}     
\fi%

\graphicspath{{figures/}{pictures/}{images/}{./}} 

\usepackage{microtype}                 
\PassOptionsToPackage{warn}{textcomp}  
\usepackage{textcomp}                  
\usepackage{mathptmx}                  
\usepackage{times}                     
\usepackage{cite}                      
\usepackage{tabu}                      
\usepackage{booktabs}      
\usepackage{subfig}
\usepackage{todonotes}

\onlineid{0}

\vgtccategory{Research}

\vgtcinsertpkg

\title{[POSTER] Semantic Augmented Reality Environment with Material-Aware Physical Interactions}

\author{Long Chen\thanks{e-mail: chenl@bournemouth.ac.uk}\\ %
        \scriptsize Bournemouth University %
\and Karl Francis\thanks{e-mail:i7241745@bournemouth.ac.uk}\\ %
     \scriptsize Bournemouth University %
\and Wen Tang\thanks{e-mail:wtang@bournemouth.ac.uk} \thanks{Corresponding Author}\\ %
     \scriptsize Bournemouth University %
}

\abstract{In Augmented Reality (AR) environment, realistic interactions between the virtual and real objects play a crucial role in user experience. Much of recent advances in AR has been largely focused on developing geometry-aware environment, but little has been done in dealing with interactions at the semantic level. High-level scene understanding and semantic descriptions in AR would allow  effective design of complex applications and enhanced user experience. In this paper, we present a novel approach and a prototype system that enables the deeper understanding of semantic properties of the real world environment, so that realistic physical interactions between the real and the virtual objects can be generated. A material-aware AR environment has been created based on the deep material learning using a fully convolutional network (FCN). The state-of-the-art dense Simultaneous Localisation and Mapping (SLAM) has been used for the semantic mapping. Together with efficient accelerated 3D ray casting, natural and realistic physical interactions are generated for interactive AR games. Our approach has significant impact on the future development of advanced AR systems and applications. 
} 

\CCScatlist{ 
  \CCScat{H.5.1}{Information Interfaces And Presentation}%
{Multimedia Information Systems}{Artificial, Augmented, and Virtual Realities};
  \CCScat{I.2.10}{Artificial Intelligence}{Vision and Scene Understanding}{3D/Stereo Scene Analysis}
}

\begin{document}


\firstsection{Introduction}

\maketitle


The latest research on Simultaneous Localisation and Mapping (SLAM) has opened up a new world for AR technology development. SLAM has changed the ways that how traditional AR systems worked in terms of camera pose estimations and tracking. Furthermore, advanced SLAM systems such as KinectFusion \cite{Newcombe2011} and Dynamicfusion \cite{Newcombe2015} are able to reconstruct dense surfaces to generate geometric information of the real scene, providing structural-aware AR environment, which is a step changer in AR technology. However, although through SLAM the geometric surfaces of the scene can be efficiently recovered, individual semantic properties of various different objects of the real world remain unknown in the AR environment. 


Realistic interactions in augmented reality (AR) applications require not only structural information, but also semantic understandings of the scene. While geometric or structural information allows accurate information augmentation and placement in AR, at the interaction level, semantic knowledge will enable realistic interactions between the virtual and the real objects, for example realistic physical interactions (e.g. a virtual glass can be shunted on a real concrete floor). More importantly, using semantic scene descriptions, we can develop high-level tools for efficient design and constructions of complex AR applications, such as large scale complex AR games.  

In virtual reality (VR) applications, recent research attempts have been made towards virtual object classifications using semantic associations to describe virtual object behaviours \cite{Chevaillier2012}. Troyer \textit{et. al.} has discussed the opportunities and challenges of using the notion of \textit{conceptual modelling} for virtual reality, and pointed out that the gap between the conceptual level and the implementation level for virtual reality is too large to be bridged in one single step \cite{DeTroyer2007}. Therefore, three phases: conceptual specification phase, the mapping phase and the generation phase have been suggested. In contrast to the research in VR, little work has been presented in conceptual modelling for AR. We argue that deep understanding of the AR environment through semantic segmentation techniques is the first step towards the high-level design concept for AR. More recently, real-time 3D semantic reconstruction has been a challenging topic in robotics with many recent works being focused on object semantic labeling \cite{McCormac2016}. 

In this paper, we propose a novel material-aware semantic AR framework by constructing 3D volumetric semantics and an occupancy map of the real objects. We focus on the labeling of material properties of the real environment and the realistic physical interactions between the virtual and the real objects in AR. Staring with KinectFusion for camera pose recovery and the 3D model reconstruction, we trained a fully convolutional network (FCN) based on VGG-16 model \cite{Simonyan15} to obtain 2D semantic material segmentation maps \cite{Zhao2017} for detecting material properties of each object in the scene. The 3D point cloud of the scene is labeled with the semantic materials, so that realistic physics can be applied during interactions through real-time inference. Our proposed framework is presented through a semantic material-aware game example to demonstrate realistic physical interactions between the virtual and real objects. 


\section{Methods}
\subsection{Camera tracking and model reconstruction}
We have adapted KinectFusion as the core camera tracking system with dense 3D model reconstructions. A Kinect depth sensor has been used to fuse the data into a single global surface model while simultaneously obtaining the camera pose by using a coarse-to-fine iterative closest point (ICP) algorithm. The tracking and modeling processes consist of four steps: (i) Each pixel acquired by the depth camera is firstly transformed into the 3D space by the camera\'s intrinsic parameters and the corresponding depth value acquired by the camera; (ii) A ICP alignment algorithm is performed to estimate the camera pose between the current frame and the reconstructed model; (iii) With the available camera poses, each consecutive depth frame can be fused incrementally into one single 3D reconstruction by a volumetric truncated signed distance function (TSDF); (iv) Finally, a surface model is predicted via a ray-casting process.

\subsection{Deep learning for material recognition}
To train a neural network for material recognition, we follow the method in \cite{Zhao2017}, the VGG-16 pre-trained model for ImageNet Large-Scale Visual Recognition Challenge (ILSVRC) is used as the initial weights of our neural network \cite{Simonyan15}. We then fine-tuned the network from 1000 different classes of materials into 23 class labels as the output based on the Materials in Context Database (MINC) that contains 3 million material samples across 23 categories. However, the Convolutional Neural Network (CNN) is specifically designed for classification tasks and only produces a single classification result for a single image. We manually cast the CNN into a Fully Convolutional Network (FCN) for pixel-wise dense outputs \cite{Shelhamer2017}. By transforming the last three inner product layers into convolutional layers, the network can learn  to make dense predictions efficiently at pixel level for tasks like semantic segmentation. Finally, we have trained the FCN-32s, FCN-16s and FCN-8s consecutively using images with material labels provided in the MINC database.

\subsection{Semantic label fusion using CRF}
\begin{figure}[]
\centering
\includegraphics[width=0.45\textwidth]{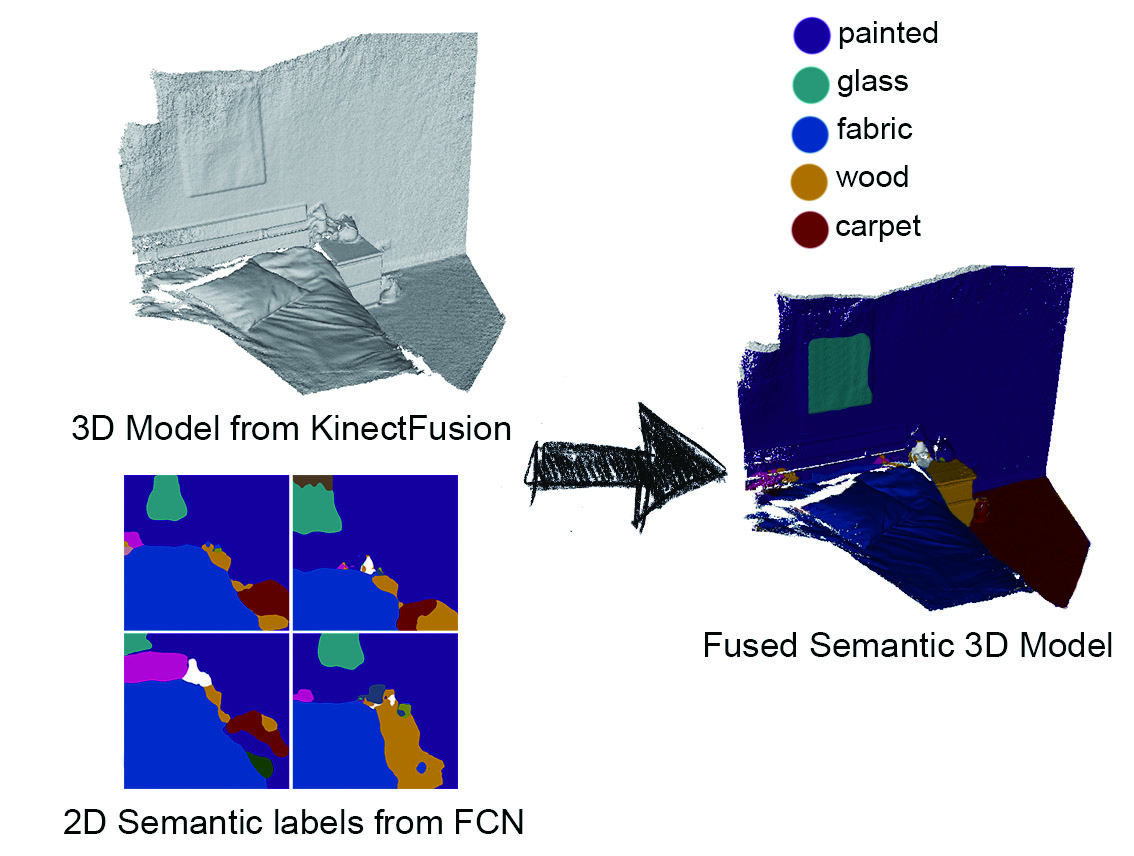}
\caption{The result of 3D model reconstruction, 3D semantic labelling and semantic 3D model fusing.}
\label{fuse}       
\end{figure}

KinectFusion builds a 3D model, but our material recognition network only provides 2D outputs. Therefore, following\cite{Armeni2016} \cite{Hermans2014} we employed a graphical model of Conditional Random Fields (CRF) \cite{Kraehenbuehl2011} to guide the fusion process of mapping the 2D semantic labels onto the 3D reconstruction model. CRF is to ensure the contextual consistency, and the final fusion result is shown in Figure \ref{fuse}.

\section{Result and discussion}
\begin{figure}[]
\centering
\subfloat[]{\includegraphics[width=0.23\textwidth]{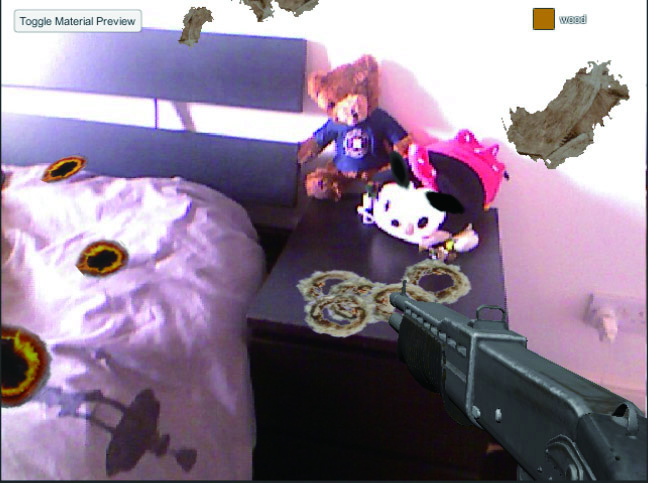}}\
\subfloat[]{\includegraphics[width=0.23\textwidth]{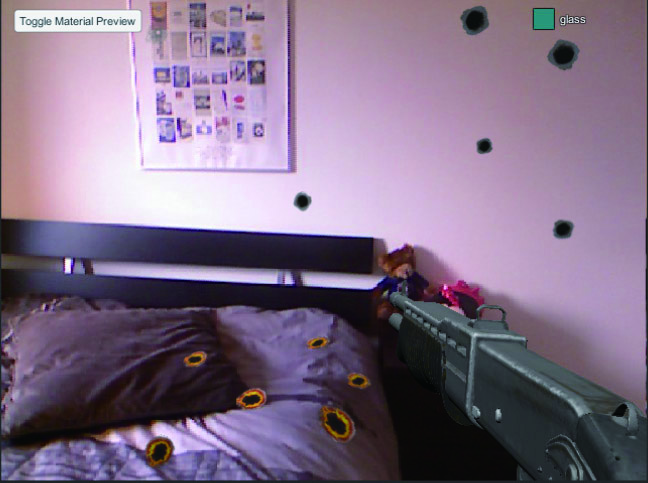}}\\
\subfloat[]{\includegraphics[width=0.23\textwidth]{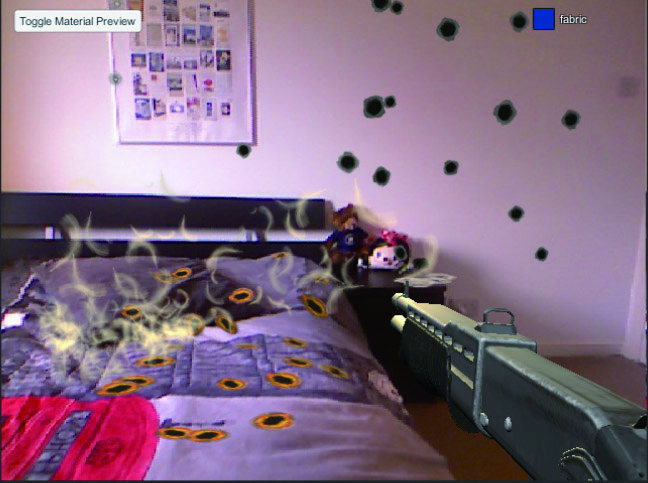}}\
\subfloat[]{\includegraphics[width=0.23\textwidth]{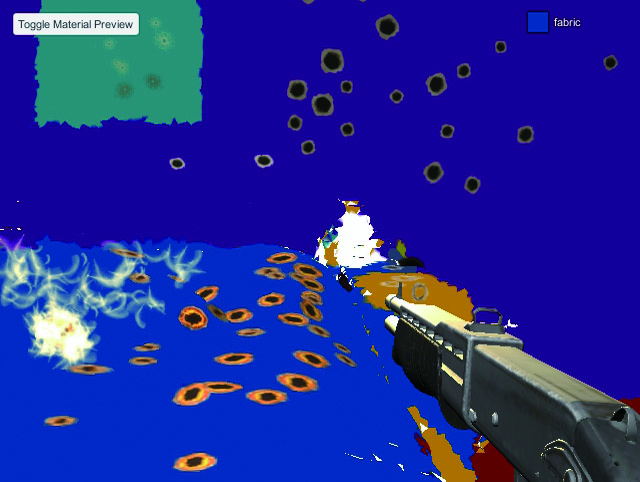}}\\
\caption{A prototype shooting game. Different material responses: (a) wood, (b) glass, (c) fabric. (d) shows the hidden material-aware layer that handles the physical interaction.}
\label{game}       
\end{figure}

We have developed a small shooting game demo \footnote{https://www.youtube.com/watch?v=02ZAqXH2FGU} (see Figure \ref{game}) in Unity to demonstrate our proposed concept of semantic material-aware AR. Our framework is built as a drop-and-play plugin in Unity, which processes the AR camera pose tracking and feeds the 3D semantic-aware model. The game contains two layers, in which the top layer displays the current video stream from a RGBD camera, whilst the semantic 3D model serves for physical interaction layer by correctly mapping the video stream with synchronised camera poses for semantic inference.  An oct-tree acceleration data structure has been  implemented for efficient ray-casting to query the material properties and corresponding physical interactions are applied through physic simulations. As can be seen from Figure \ref{game}, realistic interactions between the real and virtual objects (e.g. bullet holes, flying chips and sound) are simulated at real-time with various different material responses i.e. (a)wood, (b)glass and (c)fabric, creating a real-time interactive semantic driven AR shooting game. Our work demonstrates the first step towards the high-level conceptual interaction modelling for enhanced user experience in complex AR environment. 

\bibliographystyle{abbrv-doi}

\bibliography{template}
\end{document}